\documentclass[twoside,11pt]{article}

\usepackage{amsmath,bm,paralist}
\usepackage{jmlr2e}
\usepackage{algorithm}
\usepackage{algorithmic}

\DeclareMathOperator*{\argmin}{argmin}

\DeclareMathOperator*{\rk}{rank}
\DeclareMathOperator*{\cols}{cols}
\DeclareMathOperator*{\rows}{rows}

\def \R {\mathbb{R}}

\def \E {\mathrm{E}}

\def \e {\mathbf{e}}

\def \diag {\mbox{diag}}
\def \Gh {\widehat{G}}

\def \Wh {\widehat{W}}

\def \Uh {\widehat{U}}
\def \D {\mathcal{D}}

\def \P {\mathcal{P}}

\def \W {\mathcal{W}}

\def \Vh {\widehat{V}}

\def \Gh {\widehat{G}}
\def \Uh {\widehat{U}}
\def \Vh {\widehat{V}}
\def \Sgh {\widehat{\Sigma}}

\newtheorem{thm}{Theorem}

\begin{document}

\title{Stochastic Proximal Gradient Descent for Nuclear Norm Regularization}

\author{\name Lijun Zhang \email zhanglj@lamda.nju.edu.cn\\
       \addr National Key Laboratory for Novel Software Technology\\
       Nanjing University, Nanjing 210023, China
       \AND
       \name Tianbao Yang \email tianbao-yang@uiowa.edu\\
       \addr Department of Computer Science\\
       the University of Iowa, Iowa City, IA 52242, USA
       \AND
       \name Rong Jin \email rongjin@cse.msu.edu\\
       \addr Department of Computer Science and Engineering\\
        Michigan State University, East Lansing, MI 48824, USA
       \AND
       \name Zhi-Hua Zhou \email zhouzh@lamda.nju.edu.cn\\
       \addr National Key Laboratory for Novel Software Technology\\
       Nanjing University, Nanjing 210023, China}

\editor{Leslie Pack Kaelbling}

\maketitle

\begin{abstract}
In this paper, we utilize stochastic optimization to reduce the \emph{space} complexity of convex composite optimization with a nuclear norm regularizer, where the variable is a matrix of size $m \times n$. By constructing a low-rank estimate of the gradient, we propose an iterative algorithm based on stochastic proximal gradient descent (SPGD), and take the last iterate of SPGD as the final solution. The main advantage of the proposed algorithm is that its space complexity is $O(m+n)$, in contrast, most of previous algorithms have a $O(mn)$ space complexity. Theoretical analysis shows that it achieves  $O(\log T/\sqrt{T})$ and $O(\log T/T)$ convergence rates for general convex functions and strongly convex functions, respectively.
\end{abstract}


\section{Introduction}
Stochastic optimization has received lots of attention recently in data mining and machine learning communities~\citep{NIPS2009_AGM,COLT:Shalev:2009,ICML:13:Zhang:logT}. It is generally formulated as a minimization problem where the objective is an expectation over unknown distributions~\citep{nemirovski-2008-robust}, or defined in terms of the access model, which assumes there is a stochastic oracle that generates unbiased estimate of the objective or its gradient~\citep{COLT:Hazan:2011}. Algorithms for stochastic optimization, such as the stochastic gradient descent (SGD), utilize stochastic gradients of the objective for updating. Those methods have lightweight computation per iteration, and are widely used to reduce the \emph{time} complexity of optimization problems.

In this paper, we take a different perspective and exploit stochastic optimization as a tool to reduce the \emph{space} complexity for certain optimization problem over matrices. Specifically, we study the nuclear norm regularized convex optimization problem
\begin{equation} \label{eqn:problem:1}
\min_{W \in \W \subseteq \R^{m\times n}} \; F(W)=f(W)+\lambda \|W\|_*
\end{equation}
where $\W$ is the domain of the solution, $f(\cdot)$ is a convex function, and $\|\cdot\|_*$ is the nuclear norm of matrices. Note that the above problem is a special case of convex composite optimization~\citep{Nesterov_Composite_new}. Due to the nuclear norm regularizer, the optimal solution $W_*$ to (\ref{eqn:problem:1}) is a low-rank matrix, provided $\lambda$ is large enough. Here, we consider the large-scale setting: ``both $m$ and $n$ are very large such that directly storing an $m \times n$ matrix in memory is impossible''. This setting excludes deterministic methods for solving  (\ref{eqn:problem:1})~\citep{Nesterov_Composite_new}, as they need to evaluate the gradient of $f(\cdot)$, which needs $O(mn)$ memory. Although various stochastic algorithms has been proposed for convex composite optimization and can also be applied to (\ref{eqn:problem:1}), their primal goal is to reduce the time complexity of evaluating the gradient of $f(\cdot)$ instead of the space complexity~\citep{Sparsity_SGM,Lan:SCO}. As a result, most of them still have an $O(mn)$ space requirement.

The only \emph{memory-efficient} algorithm for solving (\ref{eqn:problem:1}) that we found in the literature is a heuristic algorithm developed by~\citet{ICML2012Avron_617}. Under the condition that it is possible to generate a stochastic gradient of $f(\cdot)$ which is also low-rank, they propose to combine SGD and truncated SVD to solve (\ref{eqn:problem:1}). By representing all the intermediate solutions and stochastic gradients in low-rank factorization forms, the space complexity can be reduced to $O(m+n)$. The major limitation of this approach is that there is no theoretical guarantee about its convergence, due to the fact that the truncated SVD operation introduces a constant error in each iteration. Furthermore, when the objective value is difficult to calculate, it is also unclear which iterate should be used as the final solution.

Inspired by the previous studies, we develop a novel stochastic algorithm that optimizes (\ref{eqn:problem:1}) in a memory-efficient way and is supported by formal theoretical guarantees. Specifically, we propose to use stochastic proximal gradient descent (SPGD) instead of SGD to solve (\ref{eqn:problem:1}), and take the last iterate of SPGD as the final solution. Similar to the heuristic algorithm of~\citet{ICML2012Avron_617}, the proposed algorithm always maintains a low-rank factorization of iterates that can be conveniently held in memory, and both the time and space complexities in each iteration are $O(m+n)$. Built upon recent progresses in stochastic optimization~\citep{ICML2012Rakhlin,ICML2013Shamir}, we further analyze the convergence property of the proposed algorithm, and show that the last iterate has an $O(\log T/\sqrt{T})$ convergence rate for general convex functions, and an $O(\log T/T)$ rate for strongly convex functions.
\section{The Algorithm}
\begin{algorithm}[t]
\caption{SPGD for nuclear norm regularization}
{\bf Input}: The number of trials $T$, and the regularization parameter $\lambda$
\begin{algorithmic}[1] \label{alg:1}
\STATE Initialize $W_1=0$
\FOR{$t = 1, 2, \ldots, T$}
\STATE Construct a low-rank stochastic gradient $\Gh_t=A_t B_t^\top$ of $f(\cdot)$ at $W_t$
\STATE Calculate $W_{t+1}$ according to (\ref{eqn:update})
\ENDFOR
\RETURN $W_{T+1}$
\end{algorithmic}
\end{algorithm}

Our algorithm is based on the following observation. For any $W \in \W$, it is always possible to generate a low-rank matrix $\Gh$ represented as $AB^\top$ such that
\[
\E[\Gh]=\E[AB^\top] \in \partial f(W)
\]
and $A$ and $B$ occupy $O(m)$ and $O(n)$ memory, respectively. The existence and construction of such a low-rank matrix can be found in previous studies~\citep{ICML2012Avron_617,Low-rank:SDP}, and will be discussed later.

Denote by $W_t$ the solution at the $t$-th iteration, and let $\Gh_t=A_t B_t^\top$ be the low-rank stochastic gradient of $f(\cdot)$ at $W_t$. Then, we update the current solution by the SPGD, which is a stochastic variant of composite gradient mapping~\cite{Nesterov_Composite_new}
\begin{equation} \label{eqn:update}
\begin{split}
&W_{t+1}\\
=& \argmin_{W \in \W} \frac{1}{2} \|W-W_t\|_F^2 +  \eta_t \langle W- W_t,  \Gh_t \rangle  + \eta_t \lambda \|W\|_* \\
= &\argmin_{W \in \W} \frac{1}{2} \left\| W - \left(W_t - \eta_t \Gh_t\right)\right\|_F^2 +  \eta_t \lambda \|W\|_* \\
\end{split}
\end{equation}
where $\eta_t >0$ is the step size. Due to the presence of the nuclear norm regularizer, all the iterates $W_t$'s tend to be low-rank matrices. The fact that both $W_t$ and $\Gh_t$ are low-rank matrices can in turn be exploited to accelerate the updating.

In the beginning, we simply set $W_1$ be the zero matrix, and take the last iterate $W_{T+1}$ as the final solution. The complete procedure is summarized in Algorithm~\ref{alg:1}.
\subsection{Construction of the Low-rank Stochastic Gradients}
For a given matrix $W$, there are various ways to construct a low-rank stochastic gradient $\Gh$ satisfying $\E[\Gh]\in \partial f(W)$~\citep{ICML2012Avron_617,Low-rank:SDP}. For brevity, we just describe a general approach. To this end, we need the following definition from~\citet{ICML2012Avron_617}.
\begin{definition}[Probing Matrix] A random $n \times k$ matrix $Y$ is a probing matrix if $E[Y Y^\top] = I$.
\end{definition}
\citet{ICML2012Avron_617} also provide several families of distributions that generate probing matrices efficiently.
\begin{lemma} Let $Y=Z/\sqrt{k}$ where $Z$ is a random matrix drawn any one of the following distributions:
\begin{enumerate}
  \item Independent entries taking values $+1$ and $-1$ with equal probability $1/2$.
  \item Independent and identically distributed standard normal entries.
  \item Each column of $Z$ is drawn uniformly at random and independent of each other from $\{\sqrt{n} \e_1,\ldots,\sqrt{n}\e_2\}$
(scaled identity vectors).
\end{enumerate}
Then, Y is a probing matrix.
\end{lemma}

Let $G$ is a subgradient of $f(\cdot)$ at $X$, i.e., $G \in \partial f(X)$. Then, we can construct $\Gh$ as
\[
\Gh=AB^\top, \ A=GY, \textrm{ and } B=Y.
\]
Thus, as long as we can evaluate the subgradient of $f(\cdot)$, we can construct a low-rank stochastic gradient. Note that when $Y$ is constructed from the scaled identity vectors, $A=GY$ are scaled columns of $G$. So, in this case, we only need to compute a small portion of $G$, which could be very efficient.
\subsection{Efficient Implementation of the Updating}
During the optimization process, all the iterates $W_t$ are represented in the SVD form $W_t= U_t^\top \Sigma_t V_t^\top$, which needs $O((m+n)r_t)$ memory. Here, $r_t$ is the rank of $W_t$. Then, by exploiting the fact that $\Gh_t=A_t B_t^\top$ is also represent in a low-rank factorization form, we can implement the updating rule in (\ref{eqn:update}) efficiently when $\W=\R^{m \times n}$ or $\W=\{W|W\in \R^{m \times n}, \|W\|_F \leq R\}$.  We start with the incremental SVD which is a core operation during the updating.

\subsubsection{Incremental SVD} \label{sec:inc:SVD}
Let $X \in \R^{m \times n}$ be a rank-$r$ matrix with economy SVD $X=U \Sigma V^\top$. Let $A=\R^{m \times c}$ and $B\in \R^{n \times c}$. Then, the economy SVD of $X+ AB^\top$ can be calculated in $O((m+n)(r+c)^2+(r+c)^3)$ time with $O((m+n)(r+c))$ memory. We provide a detailed procedure below~\citep[Section 2]{Thin:SVD}.

Let $P$ be an orthogonal basis of the column space of $(I-UU^\top)A$, and set $R_A=P^\top (I-UU^\top) A$. Note that $\cols(P) = \rows(R_A) = \rk((I-UU^\top)A) \leq c$, and may be zero if $A$ lies in the column space of $U$. Then, we have
\[
\left[
        \begin{array}{cc}
          U & A \\
        \end{array}
      \right]=\left[
        \begin{array}{cc}
          U & P \\
        \end{array}
      \right]
 \left[
              \begin{array}{cc}
                I & U^\top A \\
                0 & R_A \\
              \end{array}
            \right]
\]
Similarly, let $Q$ be an orthogonal basis of the column space of $(I-VV^\top)B$, and set $R_B=Q^\top (I-VV^\top) B$. Then, it is easy to verify
\[
X + AB^\top =\left[
        \begin{array}{cc}
          U & P \\
        \end{array}
      \right] K \left[
        \begin{array}{cc}
          V & Q \\
        \end{array}
      \right]^\top
\]
where
\[
\begin{split}
K= &\left[
              \begin{array}{cc}
                I & U^\top A \\
                0 & R_A \\
              \end{array}
            \right]\left[
              \begin{array}{cc}
                \Sigma & 0 \\
                0 & I\\
              \end{array}
            \right]\left[
              \begin{array}{cc}
                I & V^\top B \\
                0 & R_B \\
              \end{array}
            \right]^\top \\
= &  \left[
              \begin{array}{cc}
                \Sigma & 0 \\
                0 & 0 \\
              \end{array}
            \right]+  \left[
                        \begin{array}{c}
                          U^\top A \\
                          R_A \\
                        \end{array}
                      \right] \left[
                        \begin{array}{c}
                          V^\top B \\
                          R_B \\
                        \end{array}
                      \right]^\top \in \R^{(r+\cols(P)) \times  (r + \cols(Q))}
\end{split}
\]
Let the SVD of $K$ be $K=\Uh \Sgh \Vh^\top$. Then, the SVD of $X+ AB^\top$  is given by
\[
X + AB^\top =\left(\left[
        \begin{array}{cc}
          U & P \\
        \end{array}
      \right] \Uh \right)  \Sgh  \left( \left[
        \begin{array}{cc}
          V & Q \\
        \end{array}
      \right] \Vh \right)^\top
\]

\subsubsection{Updating for Unbounded Domain} \label{sec:unbounded}
We first introduce the Singular Value Shrinkage (SVS) operator with threshold $\lambda$, which is denoted by $\D_\lambda[\cdot]$~\citep{MC:SVT}. For a matrix $Y \in \R^{m \times n}$ with singular value decomposition $U \Sigma V^\top$, where $\Sigma=\diag[\sigma_1,\ldots, \sigma_{\min(m,n)}]$, we have
\[
\D_\lambda[Y]= U \D_\lambda[\Sigma] V^\top
\]
where
\[
\D_\lambda[\Sigma]=\diag\left[\max(0,\sigma_1-\lambda),\ldots, \max(0,\sigma_{\min(m,n)}-\lambda)\right].
\]
The following theorem shows that $\D_\lambda[Y]$ is the optimal solution to a unconstrained least square problem with nuclear norm regularization~\citep[Theorem 2.1]{MC:SVT}.
\begin{thm}\label{thm:svs} For each $\lambda \geq 0$ and $Y \in \R^{m \times n}$, the SVS operator obeys
\begin{equation} \label{eqn:unconstrain:svs}
\D_\lambda[Y]=\argmin_{X} \frac{1}{2} \|X-Y\|_F^2+ \lambda \|X\|_*
\end{equation}
\end{thm}

When $\W=\R^{m\times n}$, the above theorem implies
\[
W_{t+1}=\D_{\lambda \eta_t}[W_t - \eta_t \Gh_t]
\]
which can be implemented in the following two steps:
\begin{enumerate}
  \item  Since $W_t= U_t^\top \Sigma_t V_t^\top$ and $\Gh_t=A_t B_t^\top$, we can use the incremental SVD algorithm in Section~\ref{sec:inc:SVD} to find the SVD of $\Wh_{t+1}:=W_t - \eta_t \Gh_t$, which is denoted by $\Uh_{t+1} \Sgh_{t+1} \Vh_{t+1}^\top$. Let $r_t$ and $c_t$ be the rank of $W_t$ and $\Gh_t$, respectively. The time and space complexities of this step are $O((m+n)(r_t+c_t)^2+(r_t+c_t)^3)$ and $O((m+n)(r_t+c_t))$, respectively.
  \item Based on the SVD factorization $\Wh_{t+1}=\Uh_{t+1} \Sgh_{t+1} \Vh_{t+1}^\top$, $W_{t+1}=\D_{\lambda \eta_t}[\Wh_{t+1}]$ can be calculated directly according to the definition of SVS operator. Furthermore, $W_{t+1}$ is also represented in the SVD form as $W_{t+1}= U_{t+1}^\top \Sigma_{t+1} V_{t+1}^\top$. The time complexity is $O(r_t+c_t)$, and there is no additional space requirement.
\end{enumerate}

\subsubsection{Updating for Frobenius Norm Ball}
To facilitate presentations, we introduce the metric projection operator onto the Frobenius norm ball with radius $R$:
\[
\P_{R}[Y]=\argmin_{\|X\|_F \leq R} \|X-Y\|_F=\left\{
                                              \begin{array}{ll}
                                                Y, & \textrm{if } \|Y\|_F \leq R\hbox{;}\\
                                                \frac{R}{\|Y\|_F} Y  , & \hbox{otherwise.}
                                              \end{array}
                                            \right.
\]

To derive the updating rule when $\W=\{W|W\in \R^{m \times n}, \|W\|_F \leq R\}$, we need the following theorem.
\begin{thm} \label{thm:svs:fro} The optimal solution $X_*$ to the following optimization problem
\begin{equation}\label{eqn:opt1}
\min_{\|X\|_F\leq R} \;  \frac{1}{2} \|X-Y\|_F^2 + \lambda \|X\|_*
\end{equation}
is given by
\[
X_*= \P_{R}\left[ \D_\lambda[Y]\right]=\left\{
                                              \begin{array}{ll}
                                                \D_\lambda[Y], & \textrm{if } \|\D_\lambda[Y]\|_F \leq R \hbox{;}\\
                                                \frac{R}{\|\D_\lambda[Y]\|_F} \D_\lambda[Y]  , & \hbox{otherwise.}
                                              \end{array}
                                            \right.
\]
\end{thm}
From Theorem~\ref{thm:svs:fro}, we have
\[
W_{t+1}=\P_{R}\left[ \D_{\lambda \eta_t}[W_t - \eta_t \Gh_t]\right]=\left\{
                                              \begin{array}{ll}
                                                \D_{\lambda \eta_t}[W_t - \eta_t \Gh_t], & \textrm{if } \|\D_{\lambda \eta_t}[W_t - \eta_t \Gh_t]\|_F \leq R \hbox{;}\\
                                                \frac{R}{\|\D_{\lambda \eta_t}[W_t - \eta_t \Gh_t]\|_F} \D_{\lambda \eta_t}[W_t - \eta_t \Gh_t]  , & \hbox{otherwise.}
                                              \end{array}
                                            \right.
\]
To calculate $W_{t+1}$, we first use the two steps in Section~\ref{sec:unbounded} to get $\D_{\lambda \eta_t}[W_t - \eta_t \Gh_t]=U_{t+1}^\top \Sigma_{t+1} V_{t+1}^\top$, and then use the following additional step to normalize the diagnal matrix  $\Sigma_{t+1}$:
\[
\Sigma_{t+1}\leftarrow \min \left( 1, \frac{R}{\|\Sigma_{t+1}\|_F}  \right) \Sigma_{t+1}
\]
which has $O(r_{t+1})$ time complexity.

\section{Analysis}
The convergence of the last iterate of SGD has been analyzed by~\citet{ICML2012Rakhlin} and~\citet{ICML2013Shamir}. By extending their analysis, we provide the convergence of SPGD for solving the nuclear norm regularized problem in (\ref{eqn:problem:1}).

The following theorem shows that for general convex functions, the last iterate attends an $O(\log T/\sqrt{T})$ convergence rate.
\begin{thm} \label{thm:general:conver} Assume there exists constants $G$ and $D$ such that $\E [\|\Gh_t \|_F^2] \leq G^2$ for all $t$ and $\sup_{W,W' \in \W} \|W-W'\|_F\leq D$. Setting $\eta_t=c/\sqrt{T}$, we have
\[
\E\left[ F(W_T) - F(W_*)\right] \leq \left( \frac{D^2}{c} + c \left(G^2+  16r \lambda^2+ 4 \lambda G \sqrt{r} \right)\right) \frac{2+\log T}{\sqrt{T}}
\]
where $W_*$ is the optimal solution to (\ref{eqn:problem:1}) and $r \geq \max_t \rk(W_t)$.
\end{thm}

Next, we study the case that the $f(\cdot)$ is strongly convex. In this case, the optimal solution $W_*$ is unique, which allows us to bound the difference between $W_t$ and $W_*$ in each iteration.
\begin{lemma} \label{lem:strongly:convex1} Suppose $f(\cdot)$ is $\mu$-strongly convex, and $\E [\|\Gh_t \|_F^2] \leq G^2$ for all $t$. Setting $\eta_t=1/(\mu t)$, we have
\[
\E \left[ \|W_t-W_*\|_F^2\right] \leq \frac{4}{\mu^2 t} \left(G^2+  16r \lambda^2+ 4 \lambda G \sqrt{r} \right)
\]
where $W_*$ is the optimal solution to (\ref{eqn:problem:1}) and $r \geq \max_t \rk(W_t)$.
\end{lemma}
As indicated by the above lemma, the convergence rate in terms of the squared distance between $W_T$ and $W_*$ is $O(1/T)$.

Based on Lemma~\ref{lem:strongly:convex1}, it is also possible to characterizes the performance in terms of the objective function.
\begin{thm} \label{thm:strongly:convex1} Suppose $f(\cdot)$ is $\mu$-strongly convex, and $\E [\|\Gh_t \|_F^2] \leq G^2$ for all $t$. Setting $\eta_t=1/(\mu t)$, we have
\[
\E\left[ F(W_T) - F(W_*)\right] \leq 17 \left(G^2+  16r \lambda^2+ 4 \lambda G \sqrt{r} \right) \frac{1+\log T}{\mu T}
\]
where $W_*$ is the optimal solution to (\ref{eqn:problem:1}) and $r \geq \max_t \rk(W_t)$.
\end{thm}

\subsection{Proof of Theorem~\ref{thm:general:conver}}
The proof is an extension of Theorem 2 in \citet{ICML2013Shamir}, which establishes a similar guarantee for the last iterate of SGD.

From the property of strongly convex, i.e., (2) of~\citet{COLT:Hazan:2011},  the updating rule in (\ref{eqn:update}) implies
\[
\begin{split}
 & \frac{1}{2} \|W_{t+1}-W_t\|_F^2 +  \eta_t \langle W_{t+1}- W_t,  \Gh_t \rangle  + \eta_t \lambda \|W_{t+1}\|_* \\
 \leq &   \frac{1}{2} \|W-W_t\|_F^2 +  \eta_t \langle W- W_t,  \Gh_t \rangle  + \eta_t \lambda \|W\|_* - \frac{1}{2} \|W-W_{t+1}\|_F^2
\end{split}
\]
for all $W \in \W$. Then, we have
\begin{equation} \label{eqn:opt:1}
\begin{split}
 & \frac{1}{2} \|W-W_{t+1}\|_F^2+\eta_t \lambda \|W_{t+1}\|_* \\
 \leq &   \frac{1}{2} \|W-W_t\|_F^2 -\frac{1}{2} \|W_{t+1}-W_t\|_F^2+  \eta_t \langle W-W_{t+1},  \Gh_t \rangle  + \eta_t \lambda \|W\|_* \\
=& \frac{1}{2} \|W-W_t\|_F^2 +  \eta_t \langle W-W_t,  \Gh_t \rangle +\eta_t \langle W_t-W_{t+1},  \Gh_t \rangle-\frac{1}{2} \|W_{t+1}-W_t\|_F^2 + \eta_t \lambda \|W\|_*
\end{split}
\end{equation}
for all $W \in \W$.  Combining (\ref{eqn:opt:1}) with the following inequality
\[
\eta_t \langle W_t-W_{t+1},  \Gh_t \rangle-\frac{1}{2} \|W_{t+1}-W_t\|_F^2 \leq \max_W \eta_t \langle W,  \Gh_t \rangle-\frac{1}{2} \|W\|_F^2=\frac{\eta_t^2}{2}\| \Gh_t\|_F^2
\]
we have
\begin{equation} \label{eqn:opt:2}
\begin{split}
 & \frac{1}{2} \|W-W_{t+1}\|_F^2+\eta_t \lambda \|W_{t+1}\|_* \\
\leq & \frac{1}{2} \|W-W_t\|_F^2 +  \eta_t \langle W-W_t,  \Gh_t \rangle +\frac{\eta_t^2}{2}\| \Gh_t\|_F^2+ \eta_t \lambda \|W\|_*
\end{split}
\end{equation}
for all $W \in \W$.

Since $\E[\Gh_t] \in \partial f(W_t)$, by convexity, we have
\[
\begin{split}
& f(W_t)  +\lambda \|W_{t+1}\|_*- f(W) -  \lambda \|W\|_*\\
\leq & \langle W_t -W, \E[\Gh_t] \rangle -  \lambda \|W\|_* + \lambda \|W_{t+1}\|_*\\
= & \langle W_t -W, \Gh_t  \rangle +\langle W_t -W, \E[\Gh_t] -\Gh_t  \rangle -  \lambda \|W\|_* + \lambda \|W_{t+1}\|_*\\
\overset{\text{(\ref{eqn:opt:2})}}{\leq} & \frac{1}{2\eta_t} \|W-W_t\|_F^2-\frac{1}{2\eta_t} \|W-W_{t+1}\|_F^2+\frac{\eta_t}{2}\| \Gh_t\|_F^2+\langle W_t -W, \E[\Gh_t] -\Gh_t  \rangle
\end{split}
\]
and thus
\begin{equation} \label{eqn:opt:5}
\begin{split}
& f(W_t)  +\lambda \|W_t\|_* - f(W) -  \lambda \|W\|_*\\
\leq &\frac{1}{2\eta_t} \|W-W_t\|_F^2-\frac{1}{2\eta_t} \|W-W_{t+1}\|_F^2+\frac{\eta_t}{2}\| \Gh_t\|_F^2+\langle W_t -W, \E[\Gh_t] -\Gh_t  \rangle\\
&+\lambda \|W_t-W_{t+1}\|_*
\end{split}
\end{equation}
for all $W \in \W$.

Next, we discuss how to bound $\|W_t-W_{t+1}\|_*$. Choosing $W=W_t$ in (\ref{eqn:opt:2}), we have
\begin{equation} \label{eqn:opt:3}
 \frac{1}{2} \|W_t-W_{t+1}\|_F^2+\eta_t \lambda \|W_{t+1}\|_*  \leq \frac{\eta_t^2}{2}\| \Gh_t\|_F^2+ \eta_t \lambda \|W_t\|_*
\end{equation}
Let $r \geq \max_t \rk(W_t)$, we have $\|W_t-W_{t+1}\|_* \leq \sqrt{2r} \|W_t-W_{t+1}\|_F$. As a result, we have
\begin{equation} \label{eqn:opt:4}
\begin{split}
 &\|W_t-W_{t+1}\|_*^2 \leq 2 r \|W_t-W_{t+1}\|_F^2 \\
 \overset{\text{(\ref{eqn:opt:3})}}{\leq} & 4 r \left(\frac{\eta_t^2}{2}\| \Gh_t\|_F^2+ \eta_t \lambda \|W_t\|_*-\eta_t \lambda \|W_{t+1}\|_* \right) \leq 2 r \eta_t^2\| \Gh_t\|_F^2+ 4r \eta_t \lambda \|W_t-W_{t+1}\|_*
\end{split}
\end{equation}
Recall that for $x, b, c \geq 0$
\[
x^2 \leq bx +c \Rightarrow x \leq 2b + \sqrt{2c}.
\]
From (\ref{eqn:opt:4}), we have
\begin{equation} \label{eqn:opt:6}
\|W_t-W_{t+1}\|_* \leq 8r \eta_t \lambda + 2  \eta_t \| \Gh_t\|_F \sqrt{r}
\end{equation}
Combining (\ref{eqn:opt:5}) and (\ref{eqn:opt:6}), we have
\begin{equation} \label{eqn:opt:7}
\begin{split}
& f(W_t)  +\lambda \|W_t\|_* - f(W) -  \lambda \|W\|_*\\
\leq &\frac{1}{2\eta_t} \|W-W_t\|_F^2-\frac{1}{2\eta_t} \|W-W_{t+1}\|_F^2+\langle W_t -W, \E[\Gh_t] -\Gh_t  \rangle \\
 &+  \left(\| \Gh_t\|_F^2+  16r \lambda^2+ 4 \lambda \| \Gh_t\|_F \sqrt{r} \right) \frac{\eta_t}{2}
\end{split}
\end{equation}

By Jensen's inequality, we have
\[
(\E[\|\Gh_t \|_F] )^2 \leq \E [\|\Gh_t \|_F^2] \leq G^2 \Rightarrow \E[\|\Gh_t \|_F] \leq G.
\]
Taking expectation over both sides of (\ref{eqn:opt:7}), we have
\begin{equation} \label{eqn:opt:8}
\begin{split}
& \E\left[ F(W_t) - F(W) \right] \\
\leq & \frac{1}{2\eta_t} \E\left[\|W-W_t\|_F^2\right]-\frac{1}{2\eta_t} \E \left[\|W-W_{t+1}\|_F^2\right]+ \frac{\eta_t}{2}  \left(G^2+  16r \lambda^2+ 4 \lambda G \sqrt{r} \right)\\
&+\E \left[\langle W_t -W, \E[\Gh_t] -\Gh_t  \rangle \right]
\end{split}
\end{equation}
Choosing $W=W_{T-k}$, for any $t \geq T-k$, we have
\begin{equation} \label{eqn:opt:8}
\begin{split}
& \E\left[ F(W_t) - F(W_{T-k}) \right] \\
\leq & \frac{1}{2\eta_t} \E\left[\|W_{T-k}-W_t\|_F^2\right]-\frac{1}{2\eta_t} \E \left[\|W_{T-k}-W_{t+1}\|_F^2\right]+ \frac{\eta_t}{2}  \left(G^2+  16r \lambda^2+ 4 \lambda G \sqrt{r} \right)
\end{split}
\end{equation}
where we use the fact
\[
\E \left[\langle W_t -W_{T-k}, \E[\Gh_t] -\Gh_t  \rangle \right]=0, \ \forall t \geq T-k.
\]
Summing over $t=T-k, \ldots , T$, and rearranging, we get
\begin{equation} \label{eqn:opt:9}
\begin{split}
& \E\left[ \sum_{t=T-k}^T F(W_t) - F(W_{T-k}) \right] \\
\leq &\sum_{t=T-k+1}^T  \frac{\E \left[\|W_{T-k}-W_{t}\|_F^2\right]}{2} \left( \frac{1}{\eta_t} -\frac{1}{\eta_{t-1}} \right)+ \frac{1}{2}  \left(G^2+  16r \lambda^2+ 4 \lambda G \sqrt{r} \right)\sum_{t=T-k}^T \eta_t\\
\leq &\sum_{t=T-k+1}^T  \frac{D^2}{2} \left( \frac{1}{\eta_t} -\frac{1}{\eta_{t-1}} \right)+ \frac{1}{2}  \left(G^2+  16r \lambda^2+ 4 \lambda G \sqrt{r} \right)\sum_{t=T-k}^T \eta_t \\
=&  \sum_{t=T-k+1}^T  \frac{D^2}{2c} \left( \sqrt{t} -\sqrt{t-1} \right)+ \frac{1}{2}  \left(G^2+  16r \lambda^2+ 4 \lambda G \sqrt{r} \right)\sum_{t=T-k}^T \frac{c}{\sqrt{t}} \\
\leq &   \frac{D^2}{2c} \left( \sqrt{T} -\sqrt{T-k} \right)+ c \left(G^2+  16r \lambda^2+ 4 \lambda G \sqrt{r} \right) \left( \sqrt{T}-\sqrt{T-k-k}\right) \\
\leq &   \left( \frac{D^2}{2c} + c \left(G^2+  16r \lambda^2+ 4 \lambda G \sqrt{r} \right)\right) \left( \sqrt{T}-\sqrt{T-k-k}\right) \\
\leq & \left( \frac{D^2}{2c} + c \left(G^2+  16r \lambda^2+ 4 \lambda G \sqrt{r} \right)\right) \frac{k+1}{\sqrt{T}} \\
\end{split}
\end{equation}
where in the fifth line we use the inequality $\sum_{t=T-k}^T \frac{1}{\sqrt{t}} \leq 2(\sqrt{T}-\sqrt{T-k-1})$.

Then, it is straightforward to prove this theorem by following the arguments of Theorem 2 in \citet{ICML2013Shamir}. Specifically, we just need to replace (5) in \citet{ICML2013Shamir} with (\ref{eqn:opt:9}) in this paper, and the rest is identical.

\subsection{Proof of Lemma~\ref{lem:strongly:convex1}}
The proof of is similar to that of Lemma 1 in~\citep{ICML2012Rakhlin}, which is devoted to analyze the behavior of SGD.

Using the fact that $f(\cdot)$ is $\mu$-strongly convex, (\ref{eqn:opt:7}) becomes
\begin{equation} \label{eqn:lem2:1}
\begin{split}
& f(W_t)  +\lambda \|W_t\|_* - f(W) -  \lambda \|W\|_*\\
\leq &\frac{1}{2} \left( \frac{1}{\eta_t} - \mu \right) \|W-W_t\|_F^2-\frac{1}{2\eta_t} \|W-W_{t+1}\|_F^2+\langle W_t -W, \E[\Gh_t] -\Gh_t  \rangle \\
 &+  \left(\| \Gh_t\|_F^2+  16r \lambda^2+ 4 \lambda \| \Gh_t\|_F \sqrt{r} \right) \frac{\eta_t}{2}
\end{split}
\end{equation}
for all $W \in \W$. On the other hand, the strongly convexity convexity also implies
\begin{equation}\label{eqn:lem2:2}
F(W_t) - F(W_*) \geq \frac{\mu}{2} \|W_t-W_*\|_F^2
\end{equation}
From (\ref{eqn:lem2:1}) and (\ref{eqn:lem2:2}), we have
\[
\begin{split}
\|W_*-W_{t+1}\|_F^2 \leq &  \left( 1 - 2\mu \eta_t \right) \|W_*-W_t\|_F^2 + 2 \eta_t \langle W_t -W_*, \E[\Gh_t] -\Gh_t  \rangle \\
& +  \left(\| \Gh_t\|_F^2+  16r \lambda^2+ 4 \lambda \| \Gh_t\|_F \sqrt{r} \right) \eta_t^2
\end{split}
\]
Taking expectation over both sides and plugging in $\eta_t=1/(\mu t)$, we have
\begin{equation} \label{eqn:lem2:3}
\E \left[ \|W_*-W_{t+1}\|_F^2  \right ] \leq \left( 1-\frac{2}{t} \right)\E \left[ \|W_*-W_{t}\|_F^2  \right ] +  \left(G^2+  16r \lambda^2+ 4 \lambda G \sqrt{r} \right) \frac{1}{\mu^2 t^2}
\end{equation}

Next, we provide an upper bound for $\|W_*-\W_1\|_F$. From strong convexity and the fact $W_1=0$, we have
\[
\begin{split}
\frac{\mu}{2} \|W_1-W_*\|_F^2 \leq &F(W_1) - F(W_*) = f(W_1) + \lambda \|W_1\|_* - f(W_*)-\lambda \|W_*\|_* \\
\leq & f(W_1)  - f(W_*) \leq \langle \E[\Gh_1 ], W_1-W_*  \rangle \leq \|\E[\Gh_1]\|_F \|W_1-W_*\|_F
\end{split}
\]
which implies
\begin{equation} \label{eqn:lem2:4}
\|W_1-W_*\|_F^2 \leq \frac{4}{\mu^2} \|\E[\Gh_1] \|_F^2  \leq \frac{4}{\mu^2} \E [ \|\Gh_1\|_F^2 ] \leq \frac{4 G^2}{\mu^2}
\end{equation}
We complete the proof by a simple induction argument based on (\ref{eqn:lem2:3}) and (\ref{eqn:lem2:4}).
\subsection{Proof of Theorem~\ref{thm:strongly:convex1}}
The proof is similar to that of Theorem 1 in \citet{ICML2013Shamir}, and thus we just show the difference.

Following the derivation of (\ref{eqn:opt:9}), for all $W \in \W$, we have
 \begin{equation}  \label{eqn:thm2:1}
\begin{split}
 \E&\left[ \sum_{t=T-k}^T F(W_t) - F(W) \right] \leq \frac{\E \left[\|W-W_{T-k}\|_F^2\right]}{2 \eta_{T-k}} \\
&+ \sum_{t=T-k+1}^T  \frac{\E \left[\|W-W_t\|_F^2\right]}{2} \left( \frac{1}{\eta_t} -\frac{1}{\eta_{t-1}} \right)+ \frac{1}{2}  \left(G^2+  16r \lambda^2+ 4 \lambda G \sqrt{r} \right)\sum_{t=T-k}^T \eta_t\\
\end{split}
\end{equation}
provided
\[
\E \left[\langle W_t -W, \E[\Gh_t] -\Gh_t  \rangle \right]=0, \ \forall t \geq T-k.
\]

Substituting $\eta_t=1/(\mu t)$, (\ref{eqn:thm2:1}) becomes
 \begin{equation}  \label{eqn:thm2:2}
\begin{split}
\E&\left[ \sum_{t=T-k}^T F(W_t) - F(W) \right] \leq  \frac{\mu (T-k)}{2} \E \left[\|W-W_{T-k}\|_F^2\right] \\
&+ \frac{\mu}{2} \sum_{t=T-k+1}^T  \E \left[\|W-W_{t}\|_F^2\right]  + \frac{1}{2\mu}  \left(G^2+  16r \lambda^2+ 4 \lambda G \sqrt{r} \right)\sum_{t=T-k}^T \frac{1}{t} \\
\end{split}
\end{equation}
Then, Theorem~\ref{thm:strongly:convex1} can be proved by replacing (2) in \citet{ICML2013Shamir} with (\ref{eqn:thm2:2}) in this paper.
\section{Conclusion}
In this paper, we propose an memory-efficient algorithm for solving the nuclear norm regularized problem in (\ref{eqn:problem:1}). Specifically, in each iteration it constructs a low-rank stochastic gradient and updates the intermediate iterate by SPGD. Compared to previous studies, there are two advantages of the proposed algorithm: i) it space complexity is $O(m+n)$ instead of $O(mn)$ and ii) it is equipped with a formal convergence rate.

The space complexity of our algorithm also depends on the rank of the intermediate iterates $W_t$. Although $W_t$ tends to be a low-rank matrix due to the nuclear norm regularizer, an explicit upper bound about its rank is unknown. This issue will be studied as a future work. We will also investigate whether it is possible to derive efficient updating for other domains of matrices, such as the spectral norm ball.

\appendix

\section{Proof of Theorem~\ref{thm:svs:fro}}
We consider two cases: $\|\D_\lambda[Y]\|_F \leq R$ and $\|\D_\lambda[Y]\|_F > R$.

\subsection{$\|\D_\lambda[Y]\|_F \leq R$}
From Theorem~\ref{thm:svs}, we know that $\D_\lambda[Y]$ is the optimal solution to a unconstrained optimization problem in (\ref{eqn:unconstrain:svs}). Thus, if $\D_\lambda[Y] \leq R$, it is also the optimal solution to (\ref{eqn:opt1}).

\subsection{$\|\D_\lambda[Y]\|_F > R$}
We denote the singular values of $Y$ by $\sigma_1,\sigma_2,\ldots$. From (\ref{eqn:unconstrain:svs}), we know
\begin{equation}  \label{eqn:pro:1}
\begin{split}
&\min\limits_{X} \; \frac{1}{2} \|X-Y\|_F^2+ \lambda \|X\|_* \\
 = &\frac{1}{2} \|\D_\lambda[Y]-Y\|_F^2+ \lambda \|\D_\lambda[Y]\|_* =\sum_{\sigma_i \geq \lambda} \left(\lambda \sigma_i -\frac{1}{2} \lambda^2 \right)   +\sum_{\sigma_i < \lambda} \frac{1}{2} \sigma_i^2 \\
\end{split}
\end{equation}

Following the standard analysis of convex optimization \citep{Convex-Optimization}, we introduce a dual variable $\mu$ for the constraint and obtain the Lagrange dual function
\[
\begin{split}
L(\mu)= & \min_{X} \frac{1}{2} \|X-Y\|_F^2 + \lambda \|X\|_* + \mu (\|X\|_F^2-R^2) \\
=& (1+2 \mu) \min_{X} \left( \frac{1}{2} \left\|X -\frac{1}{1+2\mu} Y \right\|_F^2 + \frac{\lambda}{1+2\mu} \|X\|_* \right) + \frac{2\mu}{2(1+2\mu)}  \|Y\|_F^2 - \mu R^2 \\
\overset{\text{(\ref{eqn:pro:1})}}{=} &\frac{1}{1+2 \mu} \left[\sum_{\sigma_i \geq \lambda} \left(\lambda \sigma_i -\frac{1}{2} \lambda^2 \right)   +\sum_{\sigma_i < \lambda} \frac{1}{2} \sigma_i^2 \right] + \frac{2\mu}{2(1+2\mu)}  \|Y\|_F^2 - \mu R^2 \\
= &\frac{1}{1+2 \mu} \left[\sum_{\sigma_i \geq \lambda} \left(\lambda \sigma_i -\frac{1}{2} \lambda^2 \right)   +\sum_{\sigma_i < \lambda} \frac{1}{2} \sigma_i^2 \right] -\frac{1}{2(1+2\mu)}  \|Y\|_F^2 - \mu R^2 + \frac{1}{2}\|Y\|_F^2 \\
=& -\frac{1}{1+2 \mu} \left[ \frac{1}{2} \sum_{\sigma_i \geq \lambda} \left( \sigma_i -\lambda\right)^2  \right] - \mu R^2 + \frac{1}{2}\|Y\|_F^2\\
=& -\frac{1}{2(1+2 \mu)} \|\D_\lambda[Y]\|_F^2 - \mu R^2 + \frac{1}{2}\|Y\|_F^2.
\end{split}
\]
As a result, the Lagrange dual problem is
\[
\max_{\mu \geq 0} \; -\frac{1}{2(1+2 \mu)} \|\D_\lambda[Y]\|_F^2 - \mu R^2,
\]
and it is easy to verify the optimal dual solution $\mu_*$ satisfies
\[
\frac{1}{1+ 2\mu_*}= \frac{R}{\|\D_\lambda[Y]\|_F}.
\]

Then, we can recover the optimal primal solution from $\mu_*$ in the following way
\[
\begin{split}
X_*= &\argmin_{X} \frac{1}{2} \|X-Y\|_F^2 + \lambda \|X\|_* + \mu_* \|X\|_F^2 \\
=& \argmin_{X} \frac{1}{2} \left\|X -\frac{1}{1+2\mu_*} Y \right\|_F^2 + \frac{\lambda}{1+2\mu_*} \|X\|_* \\
=& \D_ {\frac{\lambda}{1+2\mu_*}}\left[\frac{1}{1+2\mu_*} Y \right]=\frac{1}{1+2\mu_*}\D_\lambda[Y]=\frac{R}{\|\D_\lambda[Y]\|_F}\D_\lambda[Y].
\end{split}
\]




\vskip 0.2in
\bibliography{E:/MyPaper/ref}
\end{document}